\documentclass[preprint,12pt]{elsarticle}
\usepackage{epsfig}
\usepackage{amssymb}
\usepackage{amsthm}
\usepackage{xcolor}
\usepackage{hyperref}
\usepackage{pdflscape}
\hypersetup{
  colorlinks,
  citecolor=blue,
  citebordercolor=black}

\journal{Pattern Recognition Letters}

\begin{document}

\begin{frontmatter}

\title{Advances in Human Action Recognition: A Survey}

\author{Guangchun Cheng}
\author{Yiwen Wan}
\author{Abdullah N. Saudagar}
\author{Kamesh Namuduri}
\author{Bill P. Buckles\corref{corau}}

\cortext[corau]{Corresponding author. \textit{E-mail: Bill.Buckles@unt.edu, Tel:1-940-5654869, Fax:1-940-3698652}}

\address{Dept. of Computer Science and Engineering, University of North Texas\\
        Denton, TX 76203, USA}

\begin{abstract}
Human action recognition has been an important topic in computer vision due to its many applications such as video surveillance, human machine interaction and video retrieval. One core problem behind these applications is automatically recognizing low-level actions and high-level activities of interest. The former is usually the basis for the latter. This survey gives an overview of the most recent advances in human action recognition during the past several years, following a well-formed taxonomy proposed by a previous survey~\cite{Aggarwal2011}. From this state-of-the-art survey, researchers can view a panorama of progress in this area for future research.
\end{abstract}


\end{frontmatter}

\section{Introduction}
  \label{sec:introduction}
  Human action recognition is an active topic in the field of computer vision. This is due partially to the rapidly increasing amount of video records and the large number of potential applications based on automatic video analysis such as visual surveillance, human-machine interfaces, sports video analysis, and video retrieval. Among these applications, one of the most interesting is human action recognition especially high-level behavior recognition.

An action is a sequence of human body movements, and may involves several body parts concurrently. From the viewpoint of computer vision, the recognition of action is to match the observation (e.g. video) with previously defined patterns and then assign it a label, i.e. action type.  Depending on complexity, human activities can be categorized into four levels: gestures, actions, interactions and group activities~\cite{Aggarwal2011}, and much research follows a bottom-up construction of human activity recognition. Major components of such systems include feature extraction, action learning and classification, and action recognition and segmentation~\cite{Poppe2010}. A simple process consists of three steps, namely detection of human and/or its body parts, tracking, and then recognition using the tracking results. For instance, to recognize "shaking hands" activities, two person's arms and hands are first detected and tracked to generate a spatial-temporal description of their movement. This description is compared with existing patterns in the training data to determine the action type.

\begin{figure}[htp]
  \centering
  \includegraphics[width=0.9\linewidth]{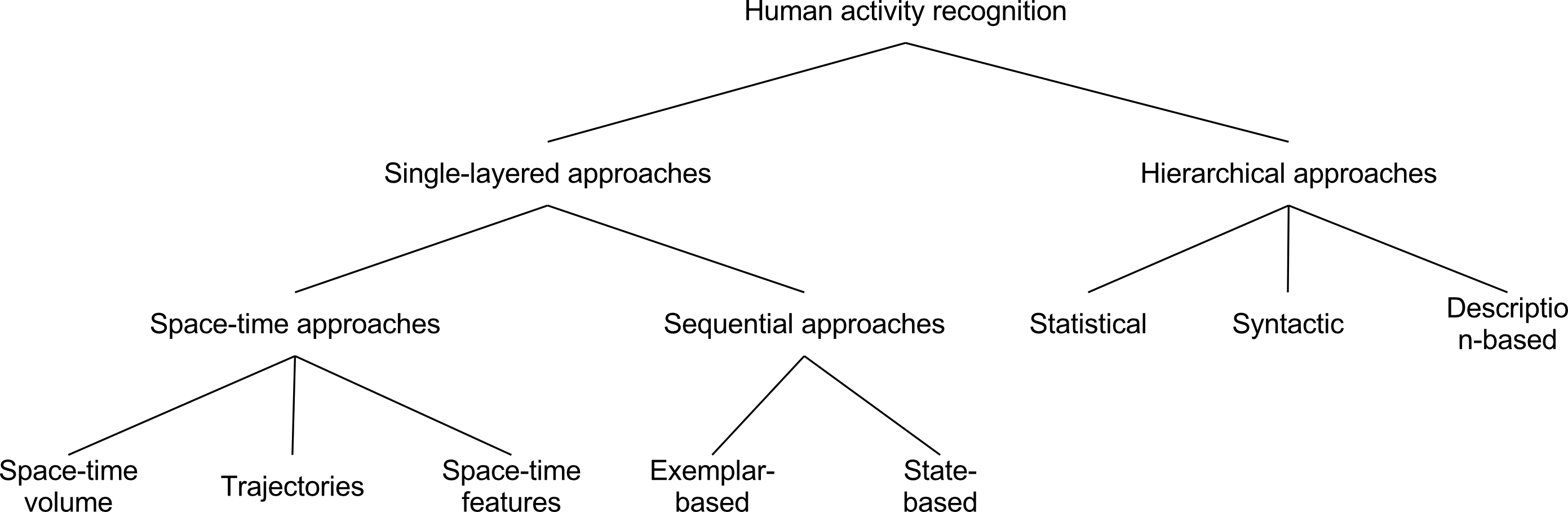}
  \caption{Hierarchical approach-based taxonomy of human activity recognition methodologies\cite{Aggarwal2011}.}
  \label{fig:hierarchicy}
\end{figure}

This paradigm heavily relies on the accuracy of tracking, which is not reliable in cluttered scenes. Many other methodologies were proposed, and can be classified according to many different criteria as in existing survey papers. Poppe~\cite{Poppe2010} discussed human action recognition from image representation and action classification separately. Weinland~\textit{et al.}\cite{Weinland2011} surveyed methods for action representation, segmentation and recognition. Turaga~\textit{et al.}\cite{Turaga2008} divided the recognition problem into action and activity according to its complexity, and classified approaches according to their ability to handle varying degrees of complexity.
There exist many other classification criteria~\cite{Aggarwal2011,Chaudhary,Algorithms2010}. Among them,Aggarwal and Ryoo~\cite{Aggarwal2011} is one of the latest comprehensive summarization and comparison of the most significant progress in this area. Based on whether the action is recognized from input images directly, Aggarwal and Ryoo~\cite{Aggarwal2011} divides the recognition methodologies into two major categories: single-layered approaches and hierarchical approaches. Both are further sub-categorized depending on the feature representation and learning methods, as shown in Fig.~\ref{fig:hierarchicy}. \cite{Aggarwal2011} surveyed progress up to three years ago.

In this paper, we focus on the state-of-the-art research not discussed in previous surveys. Additionally, in order for a comparison with previous methods, we use a similar taxonomy as in Aggarwal and Ryoo's survey\cite{Aggarwal2011}. For each of the category in Fig.~\ref{fig:hierarchicy}, recent development is presented together with the comparison between it and previously reported methods.

The remainder of this paper is structured as follows. Publicly available datasets for human action recognition are reviewed in Section~\ref{sec:datasets}, followed by two sections that review recognition approaches. In Section~\ref{sec:single-layered}, single-layered recognition approaches are reviewed with different representation and integration methods. Section~\ref{sec:hierarchical} discusses the advances in hierarchical methdologies. Section~\ref{sec:conclusion} concludes this survey.

\section{Datasets}
  \label{sec:datasets}
  In this section we discuss and describe datasets in use since 2009. Datasets that have been utilized earlier than 2009 can be found in~\cite{Aggarwal2011} in more detail. We focus on new datasets collected and we further analyze and compare them across several aspects.

\subsection{The KTH Dataset}

The current database covers six actions -- walking, jogging, running, boxing, hand waving and hand clapping performed several times by 25 subjects in four different scenarios outdoors, outdoors with scale variation, outdoors with different clothes and indoors. It contains a total of 2391 sequences. All sequences are taken with a static camera with 25fps frame rate, down sampled to the spatial resolution of 160x120 pixels. In the original paper~\cite{Sch}, sequences were divided into a training set (eight persons), a validation set (eight persons) and a test set (nine persons). The dataset does not provide background models and extracted silhouettes.

\subsection{The Weizmann Dataset}

The database covers 10 natural actions -- running, walking, skipping, jumping-jack, jumping-forward-on-two-legs, jumping-in-place-on-two-legs, galloping sideways, waving-two-hands, waving one- hand  and bending  performed by nine subjects ~\cite{ActionsAsSpaceTimeShapes_iccv05}. It contains a total of 93 sequences. All sequences are taken with a static camera with 25fps frame rate, down sampled to the spatial resolution of 180x144 pixels. The dataset also has ten additional sequences of walking captured from a different viewpoint varying between 0° and 81° relative to the image plane. The extracted masks after background subtraction and background sequences are provided.

\subsection{The IXMAS Dataset}

INRIA Xmas Motion Acquisition Sequences (IXMAS) covers 13 daily-life actions -- checking watch, crossing arms, scratching head, sitting down, getting up, turning around, walking, waving, punching, kicking, pointing, picking, overhead throwing and bottom up throwing performed three times by 11 subjects ~\cite{Weinland2006}. It contains a total of 2145 sequences.  All sequences are filmed with 5 calibrated and synchronized fire wire cameras. Dataset provides the extracted silhouettes and also reconstructed visual hulls.

\subsection{CMU MoBo Dataset}

The CMU Motion of Body (MoBo) dataset covers four different actions -- slow walking, fast walking, inclined walking, and walking with a ball -- performed by 25 subjects walking on a treadmill in the CMU 3D room~\cite{Gross_2001_3904}. More than 8000 images are captured per subject. All sequences are taken using six high resolution color cameras. The sequences are 11 seconds long at 30 fps frame rate with resolution of 640x480 pixels. The extracted silhouettes are provided.

\subsection{HOHA-1 (Hollywood Human Actions I) Dataset}

The database contains video samples covering eight actions -- answering phone, getting out a car, hand shaking, hugging, kissing, sitting down, sitting up, and standing up -- from 32 movies~\cite{laptev:08}. The two training sets are originated from 12 movies with 219 samples and test set is originated from 20 movies other than used in training with 211 samples with labels verified manually.

\subsection{HOHA-2 (Hollywood Human Actions II) Dataset}

This dataset is an extension of the HOHA dataset. The database contains video samples covering 12 actions -- answering phone, getting out a car, hand shaking, hugging, kissing, sitting down, sitting up, standing up, driving car, eating, fighting, and running -- and 10 classes of scenes from 69 movies~\cite{marszalek09}. The classes of scenes are leaving house, road and entering bedroom, car, hotel, kitchen, living room, office, restaurant, and shop. It contains a total of 3669 samples. The training set originates from 33 movies with 823 samples. The test set originates from 36 movies other than those used in training with 884 samples having labels verified manually.

\subsection{Human Eva Dataset}

The Human Eva-I dataset covers four gray scale video sequences and three color video sequences from a motion capture system which are calibrated and synchronized with 3D body poses. The database contains 4 subjects covering 6 actions -- walking, jogging, gesturing, catching, boxing and combination of walking and jogging ~\cite{LeonidSigal09}. The sequences are with resolution of 640x480 pixels captured at 60 Hz.

The Human Eva –II dataset covers extended sequence of combination of walking and jogging actions with two subjects. 

\subsection{CMU Mocap Dataset}

The CMU Mocap Dataset has six categories -- Human Interaction,  Interaction with Environment Locomotion, Physical Activities \& Sports , Situations \& Scenarios and Test Motions performed by 144 subjects. These six categories are subdivided into 23 subcategories. The actions are captured by 12 Vicon infrared MX-40 cameras with a resolution of 120 megapixel~\cite{CMUMocap06}.

Above datasets and other datasets -- UCF Sports action, UCF Youtube action and i3DPost Multi-View are summarized in Table 1

\begin{landscape}
\begin{center}
\begin{table}[ht]
\caption{Human Action Dataset}
\label{tab:1}
\begin{center}
\begin{tabular}{|p{3cm}|p{5cm}|p{2cm}|p{3cm}|p{3cm}|} 
\hline
\textbf{Dataset} &
\shortstack{\textbf{Challenges}} &
\textbf{Year} &
\textbf{Best Accuracy Achieved} &
\textbf{Category}\\ 

\hline
KTH  & Homogeneous backgrounds with a static camera & 2004 & 97.6\% [Ziaeefard \textit{et al.}'10]& General purpose action recognition \\
Weizmann & partial occlusions, non-rigid deformations, significant changes in scale and viewpoint, high irregularities in the performance of an action and low quality video & 2005 & 100\% [yangwang \textit{et al.}09; Lin \textit{et al.}09; Zeng and Ji \textit{et al.}'10] & General purpose action recognition \\
IXMAS & Multi view dataset for view-invariant human actions & 2006 & 89.4\% [Xinxiao Wu \textit{et al.}'11] & Motion Acquisition\\
CMU MoBo & Human gait & 2001 & 78.07\% [Qinfeng Shi \textit{et al.}'11] & Motion capture\\
HOHA & Unconstrained videos & 2008 & 56.8\% [Andrew Gilbert \textit{et al.}'11]& Movie\\
HOHA-2 & comprehensive benchmark for human action recognition & 2009 & 58.3\% [Heng Wang \textit{et al.}'11] & Movie\\
Human Eva & synchronized video and ground-truth 3D motion & 2009 & 84.3\% [Sang Min Yoon \textit{et al.}'10] & Pose Estimation and Motion Tracking\\
CMU MoCap & 3D marker positions and Skeleton movement & 2006 & 100\% [Hu \textit{et al.}'09] & Motion capture\\
UCF Sports & wide range of scenes and viewpoints & 2008 & 93.5\% [Simon Jones \textit{et al.}'11] & Sports action\\
UCF Youtube & Unconstrained videos & 2008 & 84.2\% [Heng Wang \textit{et al.}'11] & Sports action\\
i3DPost Multi-View & Synchronised/uncompressed-HD 8 view image sequences & 2009 & 80\% [Michael B. Holte \textit{et al.}'11] & Motion Acquisition\\
\hline
\end{tabular}
\end{center}
\end{table}
\end{center}
\end{landscape}

\section{Single-layered Approaches}
    \label{sec:single-layered}
    This section reviews the single-layered approaches. The methods are characterized by the activities to be recognized directly from the raw video data instead of primitive sub-actions or sub-activities. Therefore, most single-layered approaches deal with simple video or datasets such as KTH to recognize the actions contained. The image sequences from videos are regarded as being generated from a specific class of actions, and thus such approaches basically involve how to represent the videos (i.e. extracting features) and match them. As such, single-layered approaches mainly recognize common actions and these recognized simple primitive actions can be employed to detect more complex action recognition using hierarchical conbinations, as shown in Section~\ref{sec:hierarchical}.

As shown in a previous survey~\cite{Aggarwal2011}, various approaches have been proposed for representation and matching in single-layered systems. They can be broadly categorized into two classes: space-time approaches and sequential approaches. The core difference between space-time and sequential approaches is how the temporal dimension (i.e., the third-dimension in a 3-D XYT space) is treated. Space-time approaches treat time as a regular dimension as spatial dimensions and extract features from the 3-D volumetric videos, while sequential approaches consider a human activity as ordered observations along the timeline. Because they take sequential relationships into consideration, sequential approaches generally achieve better results than its space-time counterpart.

In this section, we present a review to the most recent progress in this branch of action recognition, and made comparison among them and previous surveyed methods. Space-time approaches are discussed in Section~\ref{sec:space-time}, and sequential approaches in Section~\ref{sec:sequential}.

\subsection{Advances in Space-Time Approaches}
  \label{sec:space-time}
For most action recognition systems (also the scope of this survey), the input is from videos. All videos discussed here consist of a temporal (T) sequence of 2-D spatial (XY) images, or equivalently a set of pixels in 3-D XYT space. Therefore, a video can be represented as a spatial-temporal volume, and this volume contains necessary information for human beings and machines to recognize the actions and activities in the volume. Based on this assumption, various representation and correspondence matching algorithms have been put forward to compactly characterize the underlying motion patterns.

As shown in Fig.~\ref{fig:hierarchicy}, we discuss the progress of space-time approaches using the same  representation-based taxonomy. Except for methods using the raw volume as a feature, all three representations use motion-related information to characterize the actions or activities.

\subsubsection{Action Recognition with Space-Time Volumes}
  \label{sec:space-time-volumes}
The most intuitive space-time volume approach would use the entire 3-D volume as feature or \textit{template}, and match unknown action videos to existing ones to obtain the classification. However, the method suffers from the noise and meaningless background information, and therefore, some effort has been made to model the foreground movement.

Based on Bobick and Davis's~\cite{Bobick01therecognition} work on movement, various approaches have been explored to extend it for action recognition. Hu \textit{et al.}~\cite{hu-action2009} proposed to combine both motion history image (MHI) and appearance information for better characterization of human actions. Two kinds of appearance-based features were proposed. The first appearance-based feature is the foreground image, obtained by background subtraction. The second is the histogram of oriented gradients feature (HOG), which characterizes the directions and magnitudes of edges and corners. SMILE-SVM (simulated annealing multiple instance learning support vector machines) was proposed for classification. It aims to obtain a global optimum via simulated annealing method without relying on model initialization to avoid local minima.

Qian \textit{et al.}~\cite{Qian2010} combined global features and local features to classify and recognize human activities. The global feature was based on binary motion energy image (MEI), and its contour coding of the motion energy image was used instead of MEI as a better global feature because it overcomes the limitation of MEI where hollows exist for parts of human blob are undetected. For local features, an object's bounding box was used. The feature points were classified using multi-class support vector machines. Roh \textit{et al.}~\cite{Roh2010} also exended Bobick and Davis's~\cite{Bobick01therecognition} MHI from 2-D to 3-D space, and proposed volume motion template for view-independent human action recognition using stereo videos.

Similarly, motivated by a gait energy image~\cite{Han:2006:IRU:1106737.1106775}, Kim \textit{et al.}~\cite{Kim2010} proposed an accumulated motion image (AMI) to represent spatiotemporal features of occurring actions. The AMI was the average of image differences. A rank matrix was obtained using ordinal measurement of AMI pixels. The distance between rank matrices of query video and candidate video was computed using \textit{L}1-norms, and the best match, spatially and temporally, was the candidate with the minimum distance.

Various researchers tried to incorporate person models such as silhouettes or skeletons for action recognition. Ikizler and Duygulu~\cite{Ikizler2009} proposed a new pose descriptor called histogram of oriented rectangles(HOR) for action recognition. They represented each human pose in an action sequence with oriented rectangular patches extracted over the human silhouette, which then formed spatial oriented histograms to represent the distribution of these rectangular patches. The local dynamics was captured with the summation of the HOR within a sliding window. Four matching methods were performed for classification, namely nearest neighbor, global histogramming, SVM and dynamic time warping.

Fang \textit{et al.}~\cite{Fang2010} first mapped the high dimensional silhouettes to low dimensional points as spatial motion description using locality preserving projection. This low-dimensional motion vector was assumed to describe the intrinsic motion structure. Then three different temporal information, i.e. temporal neighbor, motion difference and motion trajectory, was applied to the spatial descriptors to obtain the feature vectors, which were fed with \textit{k}-nearest neighborhood classifier.

Ziaeefard and Ebrahimnezhad~\cite{Ziaeefard2010} proposed the cumulative skeletonized image (CSI) across time as features, and constructed 2-D angular/distance histograms based on it. A hierarchical SVM was used for the matching process. First a coarse classification of CSI histograms using an SVM classifier was obtained with dissimilar actions, and then a second SVM was applied to confused actions using salient features among similar actions.

Wang and Mori~\cite{Wang:2009:HAR:1608576.1608765} proposed semilatent topic models (STM) following the bag-of-words framework, where a "word" corresponds to a frame and a "document" corresponds to a "video sequence". After obtaining stabilized persons in a video sequence, optical flow was computed, and half-wave rectified into four channels followed by filtering to form the motion descriptor, based on which codebook was constructed. Based on latent topics models such as LDA~\cite{Blei:2003:LDA:944919.944937} and CTM \cite{blei2006ctm}, STM does not require a choice for the number of latent topics, yet gave better training efficiency and recognition accuracy.

Guo \textit{}~\cite{Guo2009ARV} viewed an action as a temporal sequence of local shape-deformations of centroid-centered object silhouettes. Each action was represented by the empirical covariance matrix of a set of 13-dimensional normalized geometric feature vectors that captured the shape of the silhouette tunnel. The similarity of two actions was measured in terms of a Riemannian metric between their covariance matrices. The silhouette tunnel of a test video is broken into short overlapping segments and each segment was classified using a dictionary of labeled action covariance matrices and the nearest neighbor rule.

\begin{figure}[htp]
  \centering
  \includegraphics[width=0.6\linewidth]{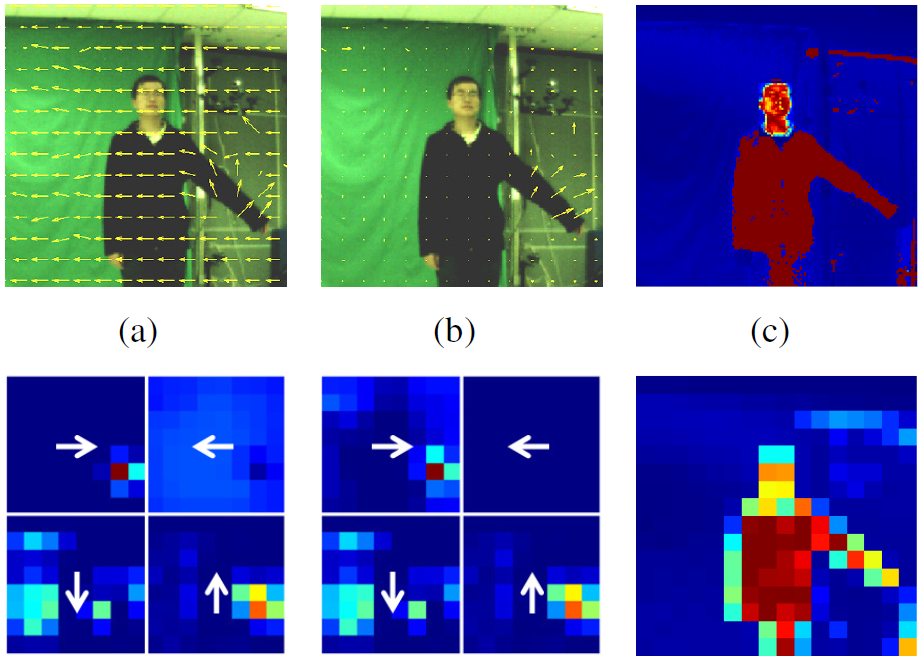}
  \caption{An example of computing the shape-motion descriptor of a gesture frame with a dynamic background from Lin \textit{et al.}~\cite{LinJD09} (\copyright 2009 IEEE). (a) Raw optical flow field, (b) Compensated optical flow field, (c) Combined, part-based appearance likelihood map, (d) Motion descriptor $D_m$ computed from the raw optical flow field, (e) Motion descriptor $D_m$ computed from the compensated optical flow field, (f) Shape descriptor $D_s$.}
  \label{fig:shape_motion}
\end{figure}


Efforts in other directions have also occurred. 
Kim and Cipolla~\cite{Kim:2009:CCA:1591906.1592380} extended Canonical Correlation Analysis (CCA) to measure video-to-video similarity. The method acted upon video volumes avoiding the difficult problems of explicit motion estimation, and provided a way of spatiotemporal matching that is robust to intraclass variations of action due to CCA. Liu \textit{et al.}~\cite{Liu2010} applied principal component analysis (PCA) to a salient action unit (SAU) (i.e., one cycle of repetitive action in a video), and AdaBoost classifier was used to classify the action in a query video. Cao \textit{et al.}~\cite{DBLP:conf/iccv/CaoLLH09} provided a new way to combine different features using a heterogeneous feature machine (HFM).

\subsubsection{Action Recognition with Space-Time Trajectories}
  \label{sec:space-time-trajectories}
Trajectory-based approaches are based on the observation that the tracking of joint positions is sufficient for humans to recognize actions~\cite{Johannson1975}. Trajectories are usually constructed by tracking joint points or other interest points on human body. Various representations and corresponding algorithms match the trajectories for action recognition.

Messing \textit{et al.}~\cite{Messing2009} extracted feature trajectories by tracking Harris3D interest points using a KLT tracker~\cite{Lucas1981}, and the trajectories were represented as sequences of log-polar quantized velocities. It used a generative mixture model to learn a velocity-history language and classified video sequences. A weighted mixture of bags of augmented trajectory sequences was modeled for action classes. These mixture components can be thought of as velocity history “words”, with each velocity history feature being generated by one mixture component, and each activity class has a distribution over these mixture components. Further, they showed how the velocity history feature can be extended, both with a more sophisticated latent velocity model, and by combining the velocity history feature with other useful information, like appearance, position, and high level semantic information.

\begin{figure}[htp]
  \centering
  \includegraphics[width=0.9\linewidth]{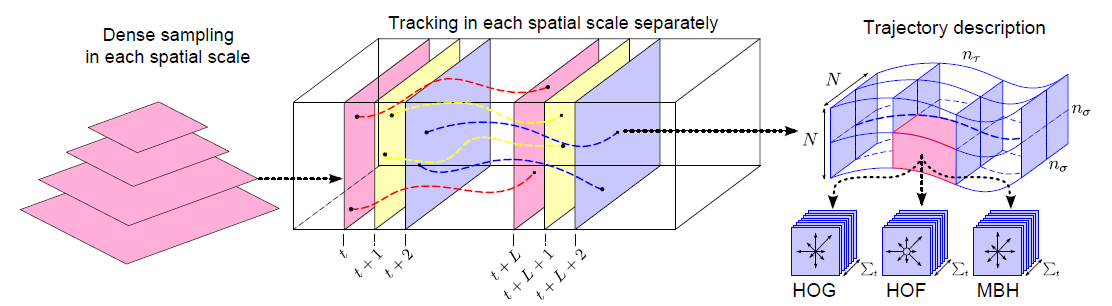}
  \caption{Illustration of dense trajectory description from \cite{wang2011} (\copyright 2011 IEEE) Left: Feature points are sampled densely for multiple spatial scales. Middle:Tracking is performed in the corresponding spatial scale over L frames. Right: Trajectory descriptors of HOG, HOF and MBH.}
  \label{fig:dense_trajectory}
\end{figure}

Wang \textit{et al.}~\cite{wang2011} proposed an approach to describe videos by dense trajectories. They sampled dense points from each frame and tracked them based on displacement information from a dense optical flow field. Local descriptors of HOG, HOF and MBH (motion boundary histogram) around interest points were computed. This is shown in Fig.~\ref{fig:dense_trajectory}.

\subsubsection{Action Recognition with Space-Time Local Features}
The application of local features in action recognition was extended from object recognition in images.  The local features refer to the description of points and their surroundings in the 3-D volumetric data with unique discriminative characteristics. These points and corresponding local feature descriptors are most informative and more robust. In terms of the density of extracted feature points, the representation of local feature approaches can be divided into two broad categories: sparse and dense. The Harris3D detector~\cite{Laptev03} and the Dollar detector~\cite{Dollar2005} are representative of the former, and optical flow-based methods the latter. Most algorithms are derived from them. Other novel methods have also been applied for finding interest points to recognize actions.

\begin{figure}[htp]
  \centering
  \includegraphics[width=0.6\linewidth]{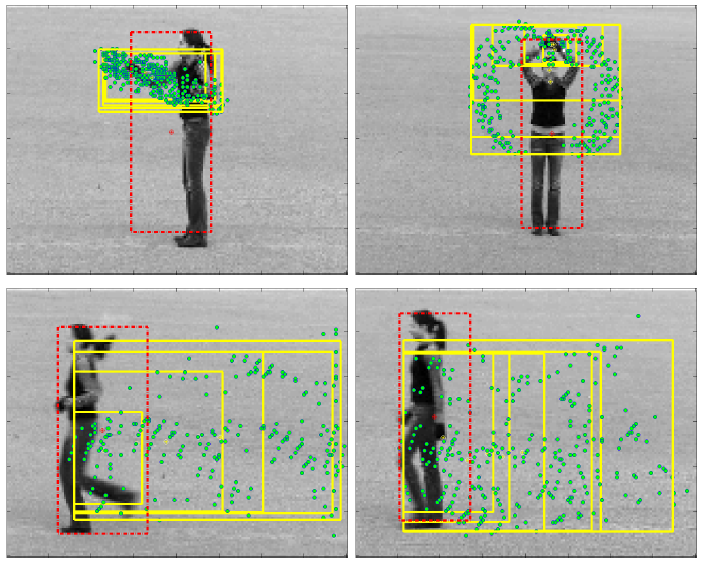}
  \caption{Examples of clouds of interest points. The clouds at different temporal scales are highlighted in yellow boxes.~\cite{Bregonzio2009} (\copyright 2009 IEEE)}
  \label{fig:clouds_interestpoints}
\end{figure}

Bregonzio \textit{et al.}~\cite{Bregonzio2009} proposed clouds of space-time interest points to overcome the limitations of the Dollar detector~\cite{Dollar2005}. Using the detected interest points from~\cite{Dollar2005}, this was achieved through extracting holistic features from clouds of interest points accumulated over multiple temporal scales followed by automatic feature selection. SVMs and Nearest Neighbor Classifiers (NCCs) were employed for classification. One example of clouds of interest points is shown in Fig.~\ref{fig:clouds_interestpoints}. Jones, \textit{et al.}~\cite{Jones2012} also based their research on the Dollar detector~\cite{Dollar2005} to detect and describe interest points which were then clustered using k-means. The innovation is that it incorporated relevance feedback mechanism by using ABRS-SVM (i.e., asymmetric bagging and random subspace support vector machine). 

In \cite{Thi2010}, space-time interest points are detected with the Harris3D detector~\cite{Laptev03}, and assigned labels of $\{-1,1\}$ indicating if it belongs to the class of interest action by using a Bayesian classifier. The feature vectors of interest point descriptors and labels are then provided to a PCA-SVM classifier to recognize the action type. In this work, the action is also localized based on CRF weighting results.

While 3D Harris corners~\cite{Laptev03} are widely used, they suffer the problem of sparity. Gilbert \textit{et al.}~\cite{Gilbert2009} used dense simple 2D Harris corners~\cite{Harris1988} in multiple scales to construct features. A two stage hierarchical grouping process was used to classify features and the actions. Sadek \textit{et al.}~\cite{Sadek11} also used a Harris corner detector in each frame and described the local feature points with temporal self-similarities defined on the fuzzy log-polar histograms. Together with global features (i.e., change of gravity centers), the feature vectors were classified with SVM.

Optical flow is also commonly used for feature point detection and description~\cite{Ikizler-Cinbis2010, Holte2011, Oikonomopoulos2009}. Ikizler-Cinbis and Sclaroff~\cite{Ikizler-Cinbis2010} employed optical flow and foreground flow to extract motion features for persons, objects and scenes, based on which the shape feature for each was also extracted. All of these feature channels were inputs to a multiple instance learning (MIL) framework to find the location of interest in a given video.

Holte \textit{et al.}~\cite{Holte2011} constructed 3D optical flow from eight weighted 2D flow fields to achieve view-invariant action recognition. 3D Motion Context (3D-MC) and Harmonic Motion Context (HMC) were used to represent the extracted 3D motion vector ﬁelds efficiently and in a view-invariant manner. The resulting 3D-MC and HMC descriptors were classiﬁed into a set of human actions using normalized correlation, taking into account the performing speed variations of different actors.

Another optical flow-based work was Oikonomopoulos's B-spline polynomial descriptor~\cite{Oikonomopoulos2009}. It was extracted as spatiotemporal salient points detected on the estimated optical flow field for a given image sequence and was based on geometrical properties of three-dimensional piecewise polynomials, namely B-splines. The latter was fitted on the spatio-temporal locations of salient points that fell within a given spatiotemporal neighborhood. The descriptor is invariant in translation and scaling in space-time.

Many efforts have been made to find interest points with other principles~\cite{Rapantzikos2009, Minhas2010, Yu2010, Shao2012, Lui2011, Zhu2009, Le2011}. For example, Rapantzikos \textit{et al.}~\cite{Rapantzikos2009} proposed a saliency-based interest points detector which incorporates intensity, color and motion. It used a multi-scale volumetric representation of the video and involved spatiotemporal operations at the voxel level. Interest points were selected as the extrema of the saliency response. Different recognition algorithms were used, such as bag-of-words with nearest neighbor for the KTH dataset and $\chi^2$ SVM kernel for HOHA dataset.

Minhas \textit{et al.}~\cite{Minhas2010} proposed new methods to compute the spatiotemporal features using 3D dual-tree discrete wavelet tranform (DT-DWT). 3D DT-DWT was employed to get the spatiotemporal information (subband vector of wavelet coefficients) efficiently, and an affine SIFT was used for local static features. By using hybrid spatiotemporal and local static features, the extreme learning machine (ELM) classifier reached high accuracy for public datasets.

Yu \textit{et al.}~\cite{Yu2010} introduced a framework based on semantic texton forests (STFs) to achieve real-time action recognition. The FAST detector~\cite{Rosten2006} was extended to V-FAST for video interest point detection. STFs are applied to classify local space-time volumes around interest points to generate the discriminative codebook. Pyramidal spatiotemporal relationship match (PSRM) was used for local appearance and structural information. A set of 3D relationship histograms were constructed by analyzing every pair of feature points using PSRM. 

Zhu \textit{et al.}~\cite{Zhu2009} proposed a new TISR (temporally integrated spatial response) descriptor, which captured the characteristics of individual actions by extracting dense spatiotemporal descriptors and representing actions by bag-of-words features. With a visual vocabulary of the TISR descriptors, the bag-of-words histogram features were able to tolerate spatial and temporal variations. 

Le ~\textit{et al.}~\cite{Le2011} presented an extension of the independent subspace anlysis (ISA) algorithm to learn invariant spatiotemporal features from unlabled video data in a hierarchical way. More specifically, features were first learnt with small input patches flattened into a vector, convolved with a larger region of the input data, and then used as input to the layer above. The features from both layers were combined as local features for classification. This two-layered stacked convolutional ISA model overcomes the limitation of ISA for large inputs, and performed well on challenging datasets.

\begin{center}
\begin{table}
\caption{Comparison of space-time approaches}
\begin{tabular}{|c|c||c|c|c|c|c|}
\hline
Approach & Category & KTH & WZMN & Other\\
\hline \hline
Hu'09 & Volume  & & & CMU:100\% \\
\hline
Ikizler'09 & Volume & 90\% & 100\% & \\
\hline
Wang'09 & Volume  & 91.2\% & 100\% & \\
\hline
Guo'09 & Volume  & 95.33\% & & \\
\hline
Kim'09 & Volume  & 95.33\% & & Gesture:82\% \\
\hline
Cao'09 & Volume & & & CMU:88.1\% \\
\hline
Liu'10 & Volume & 81.5\% & 98.3\% &\\
\hline
Ziaeefard'10 & Volume & 97.6\% & &\\
\hline
Fang'10 & Volume & & 90.21\% &\\
\hline
Qian'10 & Volume & 88.69\% & & \\
\hline
Kim'10 & Volume & 96.4\% & & \\
\hline \hline
Messing'09 & Trajectory& 89\% & & DailyAction: \\
& & &&67\% \\
\hline
Wang'11 & Trajectory & 94.2\% & &HOHA2:58.3\% \\
& & & & UCF:88.2\% \\
\hline \hline
Bregonzio'09 & Local & 93.17\% & 96.66\% &  \\
\hline
Rapantzikos'09 & Local & 88.3\% & &  \\
\hline
Minhas'10 & Local & 94.83\% & 99.44\% &  \\
\hline
Thi'10 & Local & 93.83\% & 98.2\% & HOHA:26.63\% \\
       &       &         &        & TRECVid:23.25\% \\
\hline
Ikizler-Cinbis'10 & Local &  &  & Youtube:72.51\% \\
\hline
Yu'10 & Local & 95.67\% & & UT-Itrctn:83.33\%  \\
\hline
      &       &        &  & UCF:86.5\%  \\
Le'11 & Local & 93.9\% &  & HOHA2:53.3\% \\
      &       &        &  & Youtube:75.8\% \\
\hline
Jones'12 & Local & 93.2\% & & UCF:93.5\% \\
         &       &        & & HOHA:48.4\% \\
\hline
Sadek'11 & Local & 93.6\% & 97.8\% & \\
\hline
Gilbert'09 & Local & 94.5\% & & HOHA:31.4\% \\
           &       &        & & mKTH:68.8\% \\
\hline
Oikonomopoulos & Local & 81\% & 92\% & Aerobics:95\% \\
\hline
Lui'11 & Local & 97\% & & UCF:88\% \\
\hline

\end{tabular}
\end{table}
\end{center}

\subsection{Sequential Approaches}
  \label{sec:sequential}
Single-layered sequential approaches differ with space-time approaches in that they are designed to capture temporal relationships of observations. Thus, human actions are integrated as a sequence of observations. Generally an observation is associated with local or global features extracted from a frame or a set of frames. As in ~\cite{Aggarwal2011} exemplar-based recognition and state model-based analysis are two sub-categories of sequential approaches.

\subsubsection{exemplar-based approaches}
As we mentioned earlier, sequential approaches define actions to be a sequence of observations and how observations are extracted is not limited. Exemplar-based approaches represent human actions with a template sequence of observation or a set of sample sequence of action observations. Thus the focus of exemplar-based approaches is defining how a new input video can be compared with the template or sample sequence of action observations. In previous work dynamic time warping (DTW) has been widely adopted for exemplar-based human action recognition in~\cite{darrell93,Gavrila95towards3d,Veeraraghavan_thefunction}. The similarity between input and action template is measured by comparing coefficients of the activity basis after principal component analysis (PCA) in~\cite{Yacoob1998}. Dynamic feature changes are also utilized to represent an activity as a linear-time-invariant (LTI) system~\cite{Lublinerman2006}.

Recently Lin \textit{et al.}~\cite{LinJD09} represented actions in videos as a sequence of prototypes. The prototype is based on a novel shape-motion feature and the sequence is generated by matching with a hierachical prototype tree constructed using \textit{K}-means (K=2) clustering applied iteratively. Given an action video, prototype sequence will be generated for it with a prototype sequence estimation. The prototype matching was fulfilled using FastDTW algorithm to increase computational efficiency.

\subsubsection{state model-based approaches}
Instead of representing human action as a sequence of observations state model-based approaches learn a state model for each action and each action is represented in terms of a set of hidden states. It generates sequences of observation and every sequence of observation is associated with an instance of the corresponding action. Standard hidden Markov models have been widely used for state model-based approaches in~\cite{Yamato1992Recognizing,Starner98real-timeamerican,Bobick97astate-based}. HMMs are also extended to CHSMMs to model duration of human activities~\cite{lv07,Natarajan2007}.
 
Currently, HMMs or extensions are still applied in human action recognition. In~\cite{Yu2009}, a flexible star skeleton is described for use in posture representation. The aim is to accurately match human extremities using contours and histograms from an image frame. An HMM is utilized to recognize human actions. In~\cite{Kellokumpu2009}, novel texture descriptors are proposed to describe motion and an HMM is used to model the temporal development of texture motion histograms.In ~\cite{Shi2010}, a discriminative semi-Markov model approach is proposed and in order to efficiently solve the inference problem of simultaneously segmenting and recognizing different actions they designed a Viterbi-like dynamic programming algorithm. Comparision of sequential approaches can be seen in Table ~\ref{tab:sequential}.

\begin{center}
\begin{table}
\caption{Comparison of sequential approaches}
\begin{tabular}{|c|c||c|c|c|c|c|}
\hline
Approach & Category & KTH & WZMN & Other\\
\hline \hline
Shi'11 & State-based & 95\% &  & CMU:78\%\\
       &             &      &  & WBD:94\%\\
\hline
Yu'09 & State-based &  &  & HumanClimbingFences:97.9\%\\
      &             &  &  & BalletMovie:93.6\%\\
\hline
Kellokumpu'09 & State-based & 93.8\% & 98\% & \\
\hline
Lin'09& Exemplar & 95.77\% & 100\% & \\
\hline
\end{tabular}
\label{tab:sequential}
\end{table}
\end{center}

\section{Hierarchical Approaches}
    \label{sec:hierarchical}
    As described in~\cite{Aggarwal2011} hierarchical approaches try to recognize interesting events (high-level activities) based on simpler or low-level sub-activities. In other words a high-level activity can be decomposed into a sequence of several sub-activities such as "hand shaking" may be integrated as a sequence of two hands being extended, merging into one object, and two hands being withdrawn. Sub-activities can be further considered as high-level activities until decomposed into atomic ones.

The advantage of hierarchical approaches is the capability to model the complex structure of human activities and its flexibility for either individual activities, interaction between humans and/or objects or group activities. Moreover, hierarchical models provide an intuitive and convenient interface for integrating prior knowledge and understanding of structure of activities. Hierarchical approaches to some extent have a close relationship with single-layer approaches. For example non-hierarchical single layer approaches can be easily utilized for low-level or atomic action recognition such as gesture detection. Some non-hierarchical single layer approaches can also be extended to hierarchical models such as extended multi-layered HMMs. 

Using the taxonomy proposed in~\cite{Aggarwal2011}, hierarchical approaches are categorized into three groups: statistical approaches, syntactic approaches, and description-based approaches.

\subsection{statistical approaches}
HMMs can be considered as a simple case of dynamic Bayesian networks. An HMM represents the state of the world using a single discrete random variable however DBN represents the state of the world using a set of random variables. Multiple levels of hidden states form a representation of hierarchical human activities. 
Previous research efforts on statistical approaches mainly dwell on applications of extended HMMs and dynamic Bayesian networks: 2-layered hierarchical hidden Markov models (HMMs)~\cite{Oliver02layeredrepresentations,Zhang04modelingindividual,YuAggarwal2006} and dynamic probabilistic networks (DPNs) also known as dynamic Bayesian networks (DBNs)~\cite{ShaogangGongandTaoXiang2003,Dai2008}. Sub-activities can be either concurrent or sequential. HMM-based approaches in the literature handle sequential sub-activities. Thus, a hierarchical approach using a propagation network (P-net)~\cite{Shi04propagationnetworks} has been proposed to handle both concurrent and sequential sub-activities. Beyond HMMs and DBNs a new four-layered hierarchical probabilistic latent model is proposed in~\cite{Yin2010}. First the spatial-temporal features are detected and clustered using hierarchical Bayesian model to form atomic actions. Then, based on LDA, a hierarchical probabilistic latent model is used to recognition the action without the need to specify the number of latent states. Local feature-spatial-temporal features are utilized instead of global feature such as human gesture. It is an attempt to utilize clustered space-time features as atomic actions and hierarchical descriptions and representations of complex actions. 

Another statistical approach~\cite{Han2010} is to decompose  the body into a hierarchical structure. A hierarchical manifold space is learnt to describe the motion patterns. Cascade condition random fields (CRFs) are used to predict these motion patterns. SVMs are used to classify final human actions based on the motion patterns. Hierarchical representation of human action is proposed rather than simple non-hierarchical bag-of-words representation. In~\cite{Mauthner2011} hierarchical K-means tree is also used to represent the feature cues.

The problem of insufficient training data is handled in~\cite{Zeng2010} by integrating with domain knowledge. First-order logic based domain knowledge is exploited for dynamic Bayesian network learning, both the structure and the parameters.

\subsection{syntactic approaches}
Syntactic approaches integrate actions as a string of symbols. A symbol in this context is actually the atomic sub-activities mentioned in the previous section. Atomic sub-activities can be recognized using any of the previous hierarchical or non-hierarchical techniques. However actions represented as a string of symbols results in a limitation for concurrent action recognition. In previous work context-free grammers (CFGs), based on syntactic approaches, have been studied and applied in human action recognition. Several probabilistic extension of CFGs -- stochastic context-free grammers(SCFGs) -- are introduced in~\cite{Ivanov2000,Moore02recognizingmultitasked,Minnen2003,SeongWookJoo2006}. Generally two-layer frameworks are proposed; the lower layer mostly functions to recognize atomic or low-level actions and the higher layer uses parsing techniques for the high-level activity recognition. Another limitation is that user must provide a set of production rules and in order to overcome such limitations Kitani et al.~\cite{Kitani2007} introduced an algorithm to automatically learn rules from observations.

Recently efforts have been made towards a new hierarchical framework. In~\cite{Wang2010} a four-level hierarchy is proposed. Actions are represented by a set of grammar rules categorized into three classes strong, weak, and stochastic relations based on spatio-temporal relations. 

\subsection{description-based approaches}
Description-based approaches differ from statistical and syntactic approaches through a capability to explicitly express human activities' spatio-temporal structures. Thus, such methods are able to recognize both sequential and concurrent actions instead being limited to  sequential actions. Basically, description-based approaches model human activities as an occurrence of embedded sub-activities. Such occurrences must satisfy specified temporal, spatial and logical relationships that are signatory of a high-level activity. Since the introduction of Allen's temporal interval predicates, they have been adopted for description-based human activity recognition for both sequential and concurrent relationships. Context free grammars have also been utilized for description-based approaches. A formal syntax is required for the representation of human activities as in ~\cite{Nevatia2005,Ryoo2006}. Conversion from Allen's interval algebra constraint network to a PNF-network is proposed in~\cite{Pinhanez97humanaction} to describe identical temporal information. The conversion achieves a form that is computationally tractable. Bayesian belief networks and Petri nets are introduced, respectively, in~\cite{Intille99aframework} and in~\cite{Ghanem04representationand}. Event logic is described by Siskind to recognize high-level activities in~\cite{Siskind01groundingthe}.In order to compensate for the failures of its low-level components due to the deterministic characteristics of description based approaches several probabilistic extensions of the recognition frameworks are proposed in ~\cite{Aggarwal2009,Gupta2009}.Symbolic artificial intelligence techniques Markov Logic Networks(MLN) was also adopted to infer interesting activities probabilistically as in ~\cite{Tran2008}.

Ijsselmuiden and Stiefelhagen~\cite{Ijsselmuiden2010} provide a brief framework for high-level human activity recognition. It combines different input sources and is based on temporal logic. No probabilistic computation is employed in this work.

Recently a framework was proposed in~\cite{morariu11eventstructure} to recognize behavior in one-to-one basketball by means of arbitrary trajectories obtained by tracking the ball, hands, and feet. This framework uses video analysis and mixed probabilistic and logical inference to annotate events. The method requires semantic descriptions of what generally happens in various scenarios. First-order logic based on Allen's Interval Logic is utilized to encode spatio-temporal structure knowledge and MLN is used to handle uncertainty low-level observation. 

Although, much effort has been extended as described previously but common standard dataset has not been utilized to certain extent so that comparison between description-based approaches can be expressed in terms of functionally instead of statistically. Comparison between hierarchical approaches is shown in Table~\ref{tab:hierarchical}.

\begin{center}
\begin{table}
\caption{Comparison of hierarchical approaches}
\begin{tabular}{|c|c||c|c|c|}
\hline
Approach & Category & KTH & WZMN & Other\\
\hline \hline
Yin'10 & Statistical & 82\% &  & \\
\hline
Zeng'10 & Statistical & 92.1\% & 100\% & \\
\hline
Han'10 & Statistical & & & CMU:98.27\%\\
\hline
Wang'11 & Syntactic & 92.5\% & &\\
        &           & & &HOHA:37.6\%\\
        &           & & &UCF:68.3\%\\
\hline
Ijsselmuiden'10 & Description-based & & &GroupActivities:74.4\%\\
\hline
Morariu'09 & Description-based &  &  & Basketball:72\%\\
\hline

\end{tabular}
\label{tab:hierarchical}
\end{table}
\end{center}

\section{Conclusion}
    \label{sec:conclusion}
    In this letter we provide a survey of advances in automated human action recognition. A large collection of methods are identified.  Among them, 50 specific and influential proposals of the last three years are reported. The discussion uses the same taxonomy as a previous survey based on whether the action is recognized directly from the images or low-level sub-actions. Our goal was to cover the state-of-the-art developments in each catetory, together with the datasets used in validation.

The literature reviewed shows that much research has been devoted to recognition of human actions directly from the videos or images in a single-layered manner. This is especially true for the case using space-time volume and local features. It is natural to extend 2D image processing methods, such as interest point detection, to 3D videos to extract feature descriptors. Meanwhile, more and more researchers are beginning to explore methods for high-level activity recognition. In this case, most methods surveyed use a hierarchical approach, based on statistical, syntactic, or description-based methods to explain and infer activities from low-level events. Particularly, it is of interest to combine the formal descriptors and probabilistic reasoning to interpret human actions, such as done in~\cite{Siskind01groundingthe,Nevatia2005,Ryoo2006}.

While some research has focused on complex real-world actions, most popular test datasets are still simple, constrained, and structured environments. For example, the observed actions are simple in the KTH or Weizmann datasets. Most algorithms achieve high accuracy in recognizing the actions. The introduction of more realistic datasets such as Hollywood movies and Youtube videos are challenging. The accuracy reported is low in the literature surveyed here. Based on the results of low-level actions, we hope more research will be done in the area of high-level action recognition in datasets and real-world scenes.

We know, however, that complete review of all the approaches is beyond reach. As a popular research topic, human action and activity recognition has attracted much attention and will remain important. With more and more application fields being explored, on one side, domain-specific techniques will probably emerge. On the other side, a cross-domain framework would be beneficial to the entire community.

\bibliographystyle{model2-names}
\bibliography{reference-from-mendeley}

\end{document}